\documentclass[journal]{IEEEtran}
\usepackage{xcolor,soul,framed} 

\colorlet{shadecolor}{yellow}
\usepackage[pdftex]{graphicx}
\graphicspath{{../pdf/}{../jpeg/}}
\DeclareGraphicsExtensions{.pdf,.jpeg,.png}

\usepackage[cmex10]{amsmath}
\usepackage{array}
\usepackage{mdwmath}
\usepackage{mdwtab}
\usepackage{eqparbox}
\usepackage{url}
\usepackage[hidelinks]{hyperref}
\usepackage{lmodern}
\usepackage{diagbox}
\usepackage{booktabs}

\usepackage{authblk}
\usepackage{float}

\begin{document}
\bstctlcite{IEEEexample:BSTcontrol}
    \title{Generative models-based data labeling for deep networks regression: application to seed maturity estimation from UAV multispectral images}
  \author{Eric~Dericquebourg}
  \author{Adel~Hafiane}
  \author{Raphael~Canals}

  \affil{INSA CVL, University of Orl\'eans, PRISME, EA 4229, 18022 Bourges, France}



\maketitle

\begin{abstract}
Monitoring seed maturity is an increasing challenge in agriculture due to climate change and more restrictive  practices. Seeds  monitoring in the field is essential to optimize the farming process and to guarantee yield quality through high germination. Traditional methods are based on limited sampling in the field and analysis in laboratory. Moreover, they are time consuming and only allow monitoring sub-sections of the crop field. This leads to a lack of accuracy on the condition of the crop as a whole due to intra-field heterogeneities.  Multispectral imagery by UAV allows uniform scan of fields and better capture of crop maturity information.  On the other hand, deep learning methods have shown tremendous potential in estimating agronomic parameters, especially maturity. However, they require large labeled datasets. Although large sets of aerial images are available, labeling them with ground truth is a tedious, if not impossible task. In this paper, we propose a method for estimating parsley seed maturity using multispectral UAV imagery, with a new approach for automatic data labeling.  This approach is based on parametric and non-parametric models to provide weak labels. We also consider the data acquisition protocol and the performance evaluation of the different steps of the method. Results show good performance, and the non-parametric  kernel density estimator models can improve neural network generalization when used as a labeling method, leading to more robust and better performing models. 
\end{abstract}


%
\IEEEpeerreviewmaketitle


\section{Introduction}
The increasing variability of climate change raises concerns that agricultural yields will not be able to scope with population growth in the coming years~\cite{abd-elmabod_climate_2020}. Increased temperatures associated with global warming are altering crop cycles and locations, reducing cultivable areas for some crops \cite{marklein_projected_2020}. Seed multiplication is one of the strategic agricultural sectors facing these changes. Climate impacts all stages of plant life, such as flowering, pollination, fertilization and seed filling duration. Indeed, variability of weather conditions directly influences germination and quality of seeds produced~\cite{maity_climate_2013, singh_impacts_2013}.  Seed maturity measurement is an important factor for seed quality determination.



To maintain seed compliance, several methods have been implemented for quality control. Jalink et. al. \cite{jalink_werkwijze_1997} developed a non-destructive technique for measuring seed maturity based on Laser Induced Fluorescence (LIF). This approach measures fluorescence emissions of chlorophyll $A$  at a given wavelength (around 660 nm, corresponding to the absorption peak of chlorophyll $A$). Using this procedure, chlorophyll fluorescence (CF) has been correlated with the quality of carrot, tomatoes, and cabbage seeds \cite{groot_effect_2006, jalink_chlorophyll_1998, jalink_seed_1999}. Fluorescence emissions are negatively correlated with seed chlorophyll and seed germination quality \cite{kenanoglu_chlorophyll_2013, demir_chlorophyll_2013}. CF  estimation is an area of research tackled at various levels. It ranges from portable field equipment to laboratory spectrometers and computer vision approaches \cite{ni_estimation_2019}. More specifically applied to seeds, computer vision approaches are mainly based on seed pigmentation information, geometric features \cite{antonio_computerised_2006} \cite{rodriguez-pulido_preliminary_2012} combined with infrared imaging \cite{ooms_evaluation_2011} \cite{elmasry_recent_2019}, and near infrared hyperspectral imaging \cite{wang_maturity_2021}. However, these techniques are limited to small samplings over the field and cannot reflect the heterogeneity of the field. Moreover, seed collection and preparation is time consuming. Unmanned aerial vehicles (UAVs) and multispectral imagery are emerging as an attractive approach for field-scale vegetation measurements with high spatial and temporal resolutions. 

Indeed, vegetation monitoring based on UAV imagery combined with multispectral imagery has produced notable results on various applications such as yield estimation or vegetation vigor monitoring \cite{chakraborty_assessing_2020, modica_monitoring_2020, jiang_uav-based_2021, shammi_use_2021}, and even forecasting vegetation evolution using temporal components \cite{spadoni_analysis_2020, marques_ramos_random_2020}. To extract specific information from multispectral images, vegetation indices (VIs) have been proposed, which result from arithmetic combinations of images of different wavelengths. In agriculture, they can highlight vegetation health, photosynthetic activity, and leaves chlorophyll concentration~\cite{yoder_normalized_1994, Haboudane2008, Chaoyang2008, pocas_remote_2020, cao_comparison_2019, Lang2020,  tenreiro_using_2021}. The best-known indices are the Normalized Difference Vegetation Index (NDVI)~\cite{penuelas_assessing_1993} for chlorophyll content, and the Normalized Difference Red-Edge (NDRE) for plant nitrogen diagnosis~\cite{lu_evaluating_2017}. Other indices such as the Soil Adjusted Vegetation Index (SAVI) or the Leaf Area Index (LAI)~\cite{elvidge_comparison_1995} are also suitable for plant observation. 
Statistical methods and machine learning are also significant approaches which use VIs for building models to detect crop chlorophyll content~\cite{Qiao2022}.

The past decade has seen the rapid development of deep learning approaches in many agriculture applications, such as weed management, disease detection, ... \cite{Hasan2021, Ouhami2021}. Deep learning methods have achieved state-of-the-art accuracy results \cite{abiodun_comprehensive_2019} compared to other machine learning approaches such as Support Vector Machine (SVM)~\cite{ahmed_classification_2012} or Random Forests~\cite{lottes_uav-based_2017}. Each category generally addresses different use cases: multiclass classification to discriminate portions of vegetation in images \cite{anagnostis_deep_2021, xue_fast_2019, tetila_detection_2020}, regression to estimate the evolution of a specific marker~\cite{chen_deep_2020, kim_regression_2021, lathuiliere_comprehensive_2020}, time series classification for prediction of measurements, using recurrent neural networks (RNNs)~\cite{hamrani_machine_2020, chen_research_2018}. 
Deep learning approaches have impressive capacities for data modeling, their use for chlorophyll estimation from VIs seems an interesting approach. However, the performance of deep learning approaches is strongly dependent on the amount of labeled data. Collecting samples and labeling a large amount of field data with the associated images is a very tedious and time-consuming task, if not impossible to complete.

Semi-supervised learning is one of the most widely used approaches to address the dependency on large, labeled datasets, where the goal is to combine a small labeled datased with a large set of unlabeled data~\cite{ DBLP:journals/corr/abs-2103-00550}. Semi-supervised learning deals with incomplete labeled data, while weakly supervised learning additionally deals with noisy data~\cite{sun_weakly-supervised_2020, algan_image_2021}. 
Our aim is to incorporate label uncertainty and small labeled datasets to build efficient deep network models through the concept of weakly supervised learning. However, despite the success of such a concept, way data is annotated remains an open problem that depends on the type of data and the application.

In this paper, we propose a new approach that considers generative models to associate an approximate CF value to NDVI image in order to build a large labeled dataset for deep learning models. We used both parametric and non-parametric estimation techniques, namely: gaussian mixture model (GMM), K-nearest neighborhood (KNN) and the  kernel density estimator (KDE). The first step consisted in the acquisition of aerial images, and the collection of seed samples from different locations in the crop fields. Correlation analysis between CF and NDVI was performed to identify relevant regression variables.  Then, generative models were built with few ground truth samples collected from the studied fields. The created labels  were fed into both convolutional neural networks (CNN) and long short-term memory (LSTM) to build regression models that predict chlorophyll fluorescence emissions as a function of NDVI. The proposed method proved to have a very interesting potential to deal with CF estimation from remote sensing images and to adress the problem of labeling large amounts of data.



The paper is organized as follows: the first section describes the experimental protocol and the equipment used for data acquisition, followed by the exploratory analysis of the data.  
We then present our approach and discuss results. In the last section, we conclude the paper with some perspectives.

\section{Materials and Methods}
\subsection{Data Acquisition}
The data acquisition campaign was conducted during the 2020 summer, from July to September, covering the seed filling period. Four parsley fields located in the Centre-Val de Loire region, France, were considered, representing two types of parsley, curled and flat, and four different varieties. These field choices were made to observe potential variations between both parsley types and varieties during the seed maturation phase. The mean distance between each field is about 15 km.  Data collected are of two types, ground seed samples and multispectral UAV images, both associated with a temporal dimension. As crop maturity varies rapidly in the weeks before harvest, a 3-day interval between the different data acquisitions was targeted. However, due to external factors such as weather or wind conditions preventing optimal UAV flights, the effective time step between flights varied from 3 to 6 days. This variable time interval between image acquisitions resulted in photographing the fields at different times of the day, under different weather and lighting conditions, adding natural image data variance and better representing real outdoor conditions.

\begin{figure*}[t]
\centerline{\includegraphics[width=40pc]{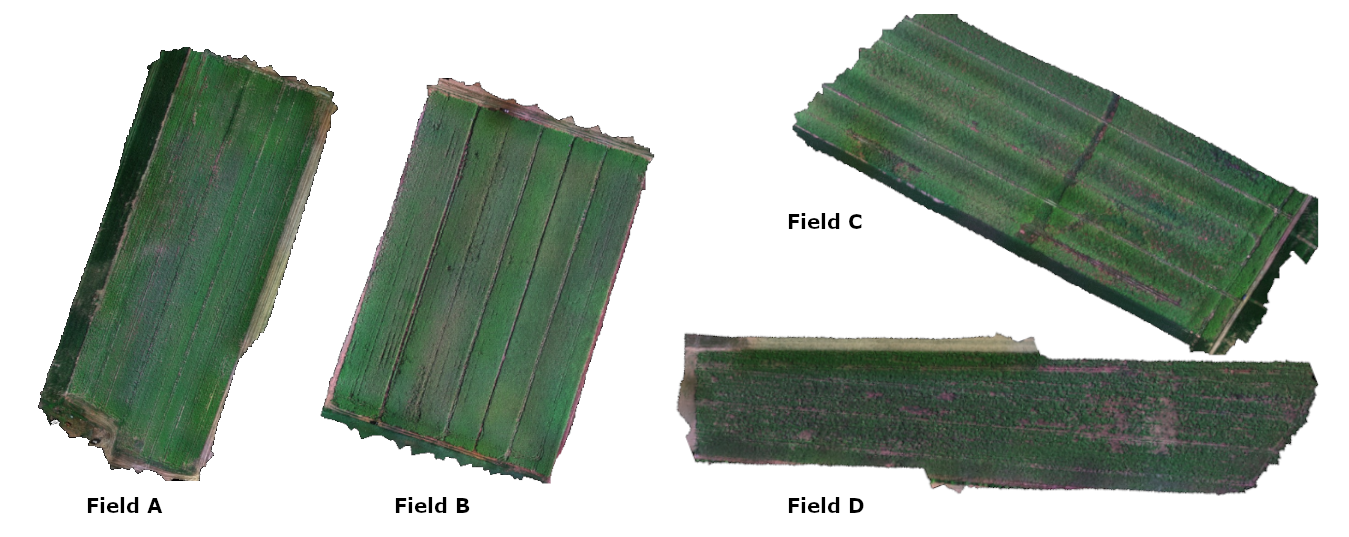}}
\caption{UAV ortho-rectified images of curled-type parsley fields A \& B and flat-type parsley fields C \& D acquired at 40 m height.}
\label{fig:fig1 fieldA and fieldB}
\end{figure*}

\subsubsection{Drone multispectral images}
Aerial images were acquired at a flight height of 40 m, a height that was chosen as a trade-off between image resolution and sufficient image ground footprint coverage. At an altitude of 40 m, the ground sample distance between two consecutive pixels is 2.73 cm and the ground footprint of a single image covers 34.9 m $\times$ 26.2 m. Therefore, each 4 ha (depicted in \autoref{fig:fig1 fieldA and fieldB}) could be covered by the UAV in approximately 20 minutes. The on-board camera used was the Micasense RedEdge-MX. 
 It is a 5-band multispectral sensor, 5 images are simultaneously acquired in different wavelengths summarized in \autoref{tab:multispectral bandwidths}.


\begin{table}[H]
 \centering{
 \caption{Multispectral specifications of the sensor}\label{tab:multispectral bandwidths}
 \begin{tabular}{|c|c|c|}
 \hline
 Band name & Wavelength center(nm) & Bandwidth(nm)\\ \hline
 Blue                		& 475 & 32 \\ \hline
 Green                    	& 560 & 27 \\ \hline
 Red		                & 668 & 14 \\ \hline
 Red-Edge		            & 717 & 12 \\ \hline
 Near-IR		            & 842 & 57 \\ \hline
 \end{tabular}
}
\end{table}
This sensor was chosen as it covers both the wavelength used by the LIF method which is centered around 660 nm (method for estimating seed maturity), and the spectral ranges corresponding to photosynthesis absorption of plants. The multispectral camera measures the reflectance of light from different types of surfaces, soil and vegetation. The values are directly influenced by external light conditions. To compare the evolution between UAV images of the same field acquired at different times, multispectral images need to be calibrated. This is done by using two external devices: the sunlight sensor which measures both the sun irradiation and the angle of incidence, and secondly, a photograph of a calibration panel whose reflectance surface is known. 
Once corrected, the multispectral images of 1280$\times$960 pixels are assembled into an ortho-rectified image, as shown in \autoref{fig:fig1 fieldA and fieldB}, using the \textit{Agisoft Metashape} software, which corrects for image distortions and GPS errors using stereo image calibration between matched points. Successive images were acquired with a minimum longitudinal and lateral overlap of 70\% and 30\% respectively, to have enough matched points in adjacent images. The overlap is usually increased depending on vegetation density.

Pixel-corrected GPS positions in the assembled images allow the same field acquired at different times during the seed filling period to be overlaid. By spatially aligning global field images at different timesteps, we can better monitor the evolution of subareas through multispectral imagery with different UAV flights.

\subsubsection{Ground Truth}
\begin{figure}
\centerline{\includegraphics[width=12pc]{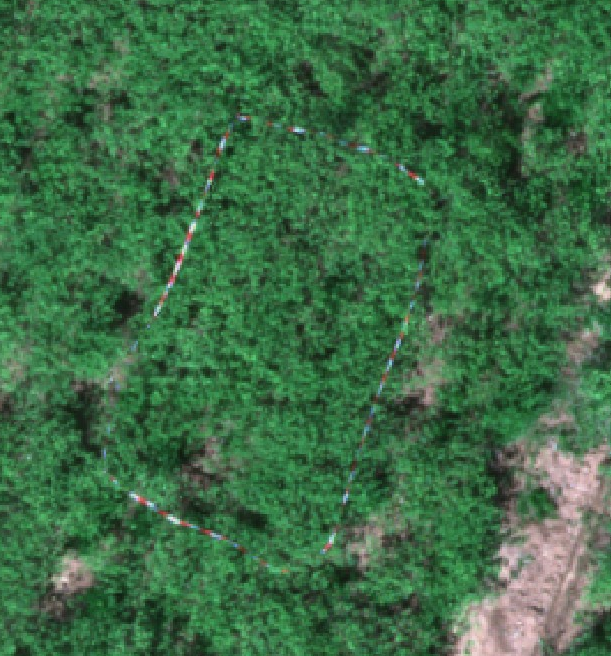}}
\caption{Example of a control zone set up in a parsley field marked with a ribbon to monitor maturity in UAV images according to ground truth harvested samples.}
\label{fig:control zone image}
\end{figure}

The ground truth, for each parsley field, is composed of the pairing between UAV images and seed samples. To guarantee a correct pairing, physical control zones were set up.~\autoref{fig:control zone image} illustrates a sampling area. For each studied parsley field, four 12 m$^{2}$ zones were marked with ribbons. This ensured that seed samples were collected from the same field subzones during the sampling period, and were easily identifiable on aerial images. The locations of the control zones were selected accross the field to represent differences in crop maturity.

Thirty-two UAV flights were carried out to monitor maturity of parsley seeds, resulting in 128 seed samples collected from the control zones. Each sample was composed of approximately 200 g of seeds, which were cleaned of waste before being processed. Maturity of seed samples is quantified by the non-destructive LIF method based on CF estimation. As plant and seed maturity increases, the CF value decreases (i.e. negative correlation)~\cite{groot_effect_2006}. The CF estimation machine provides an average CF value ranging from 0 to 10 000 pA as lower and upper theoretical bounds, which are not reached in practice. Once the CF falls below a certain threshold, the field is considered ready for swathing.
\begin{figure*}
\centerline{\includegraphics[width=45pc]{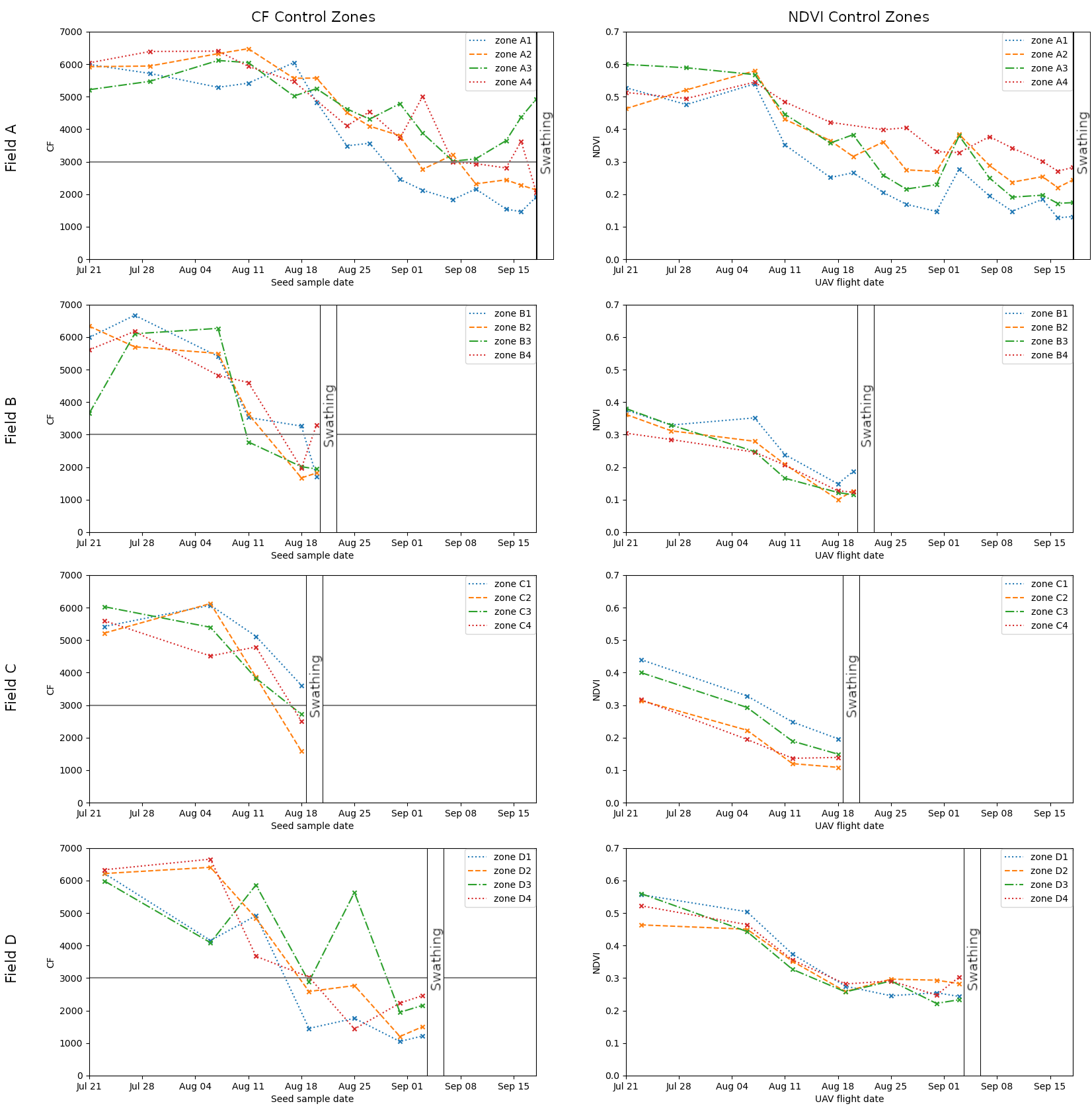}}
\caption{In the left column, the chlorophyll fluorescence (CF) evolution over time of the ground truth seed samples for fields A, B, C, D. In the right column, the associated NDVI evolutions of the control zones computed from the UAV multispectral images until the field swathing date. }
\label{fig:cf and ndvi evolution graphs}
\end{figure*}

As can be observed in~\autoref{fig:cf and ndvi evolution graphs}, the CF values of the seed samples for each control zone and each field decrease over time. 
The observation time of the four field varies because of the crop types (curled, flat) and varieties. Particular crop phenotypes of a same specie mature at distinct speeds; therefore the four fields were harvested at different times. This resulted in an unbalanced distribution of samples, with fields B and C being under-sampled compared to fields D and A.

The CF decrease is not steady, and fluctuates with punctual increase spikes, but following a global decrease. These CF fluctuations are influenced by external factors such as weather, soil type or irrigation, but also by potential sampling errors, as some control zones become heterogeneous as crops matured. The CF trend of the control zones within a same field follow similar amplitude variations and maturation durations. Greater differences were observable between fields, with early varieties having faster and steeper diminutions in CF than late ones.
With aerial images, we calculated the vegetation 
indices corresponding to the control zones and associated these indices to the estimated seed maturity of the ground samples.

\begin{figure*}
\centerline{\includegraphics[width=40pc]{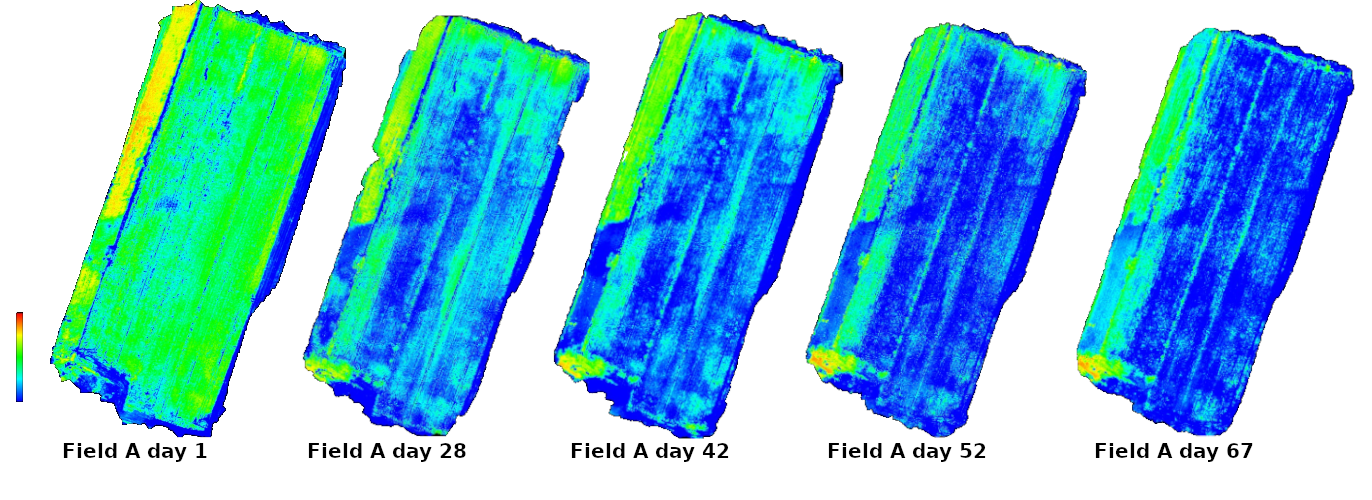}}
\caption{NDVI ortho-rectified heatmaps evolution over time. Warmer colors represent higher NDVI values, ranging from +1 to 0, as we monitor vegetation evolution.}
\label{fig:fig3 field ndvi evolution}
\end{figure*}

\subsection{Dataset}

\subsubsection{Data preparation}
Using multiple UAV flights and aligned ortho-rectified images, we can monitor the evolution of each field throughout the seed maturity period. Harvest dates were different from one field to another. We have four fields A, B, C and D, with 15, 6, 4 and 7 aerial observations respectively.  The image dataset can be divided into two categories: a small labeled set of image samples that are matched with ground truth CF values, and a large number of unlabeled samples for which we have no ground truth CF. The ground truth samples consist of 128 CF/image pairs. The remaining unlabeled dataset contains 19,443 image samples, obtained by dividing the field image into 128$\times$128 pixel tiles. This size was selected to be consistent with the control area size (i.e. resolution of 2.73 cm per pixel for an altitude of 40 m).


Since drone images of successive fields are aligned, each tile can be observed over a period, enabling the creation of a time series composed of sequential images of the same area. In this study, we limited the length of the time series to 4 consecutive observations as this is the maximum length we have for field D. For the other three fields, we selected 4 series by sliding a window, with 4 observations, over the temporal dimension. Once the 128$\times$128 images have been extracted from the ortho-photos, those that do not contain minimal 85\% of vegetation have been removed. 

\subsubsection{Correlation between CF and NDVI}
%
The Normalized Difference Vegetation Index (NDVI) is an arithmetic imagery indicator computed using images acquired in two spectral wavelengths (Red 668 nm and NearIR 842 nm) as expressed in \autoref{eq:1}. This indice is widely used for monitoring vegetation vigor and plant health from satellite and UAV imagery. NDVI values range from -1 to +1, with 0 representing no vegetation presence and values of 0.8 and above representing maximum vegetation cover.
\begin{equation} \label{eq:1}
NDVI = \dfrac{NearIR - Red}{NearIR + Red}
\end{equation}
As for the seed CF samples, the NDVI evolution in the control zones was monitored until the field harvest. The decrease in NDVI at different timesteps can be observed in \autoref{fig:fig3 field ndvi evolution}, representing the indice values using a heatmap. We observe the decrease of the indice values as the field seeds are maturing. To correlate the CF evolution with NDVI \cite{yoder_normalized_1994}, we computed the mean indice value for each control zone. The NDVI trend during the maturation period also has a decreasing trend like the CF, and shows similar variations at given timesteps, as depicted side by side in \autoref{fig:cf and ndvi evolution graphs}. The NDVI amplitudes at the beginning of the study period vary depending on the crop variety. 

The ground truths of the CF and NDVI dimensions present similar behavior when considering individual fields. This similarity does not necessarily hold for different types of fields or varieties. Assuming a direct correlation between CF and NDVI would lead to a low quality estimate as the Pearson coefficient is centered around 0.7812 when considering each field seperately and drops to 0.6460 when calculated for the four fields studied. It would probably deacrease if more plant varieties were added. Since we aim to allow for stronger correlation, we have introduced the date as an extra dimension. As can be seen in the correlation matrix transcribed in \autoref{tab:Pearson Correlation Matrix}, the couple (Date, CF) has a negative Pearson coefficient of -0.6913 and the couple (Date, NDVI) a value of -0.5267 when using all samples from the four fields. A second correlation matrix denoted in \autoref{tab:Spearman Correlation Matrix} was also calculated, using the Spearman coefficient as a non-parametric measure of rank correlation, whereas the Pearson coefficient is a parametric measure of linear correlation between variables. Both Pearson and Spearman correlation matrices output similar results. The coefficients of the different dimensions being of important values, their combination will allow a better modeling of the distribution of the data represented by the samples.
\begin{table}
\caption{Pearson Correlation Matrix}
 \centering{
 \begin{tabular}{|c|c|c|c|}
 \hline
 \textbf{Pearson} & NDVI & CF & Date    \\ \hline
 NDVI & $1$ & $0.645975$ & $-0.526745$  \\ \hline
 CF   & $0.645975$ & $1$ & $-0.691285$  \\ \hline
 Date & $-0.526745$ & $-0.691285$ & $1$ \\ \hline
 \end{tabular}
}
\label{tab:Pearson Correlation Matrix}
\end{table}
\begin{table}
\caption{Spearman Correlation Matrix}
 \centering{
 \begin{tabular}{|c|c|c|c|}
 \hline
 \textbf{Spearman} & NDVI & CF & Date    \\ \hline
 NDVI & $1$ & $0.649225$ & $-0.525424$  \\ \hline
 CF   & $0.649225$ & $1$ & $-0.715251$  \\ \hline
 Date & $-0.525424$ & $-0.715251$ & $1$ \\ \hline
 \end{tabular}
}
\label{tab:Spearman Correlation Matrix}
\end{table}
Under the assumption of correlation between the target CF and the variables (Date, NDVI), we fit the ground truth samples with generative models that will be used for the data labeling phase. The following section describes the parametric and non-parametric models used in this process.

\subsection{Labeling based on generative models}
Generative models are usually combined with neural networks~\cite{trentin_parzen_2018, li_novel_2011} for prediction or optimisation of model hyperparameters~\cite{nguyen_long-term_2020}, but not directly for data labelling. Let us now describe the considered parametric and non-parametric methods and how their respective parameters were fitted to our acquired ground truth dataset.

\subsubsection{Gaussian Mixture Model (GMM)}
GMM is a parametric method combining $n$ gaussians, where each gaussian clusters a subdivision of data. The gaussian fitting is performed by  
the iterative expectation maximization  algorithm. Combining multiple gaussians enables a better characterization of the data compared to a single gaussian, as each subdistribution is locally approximated. When data are fitted with gaussians, soft clustering is performed, since each prediction is quantified by a probability and not by a continuous target value. The probability density function of a multivariate GMM is given by~\autoref{eq:2}.
\begin{equation} \label{eq:2}
	p(\vec{x}) = \sum_{i=1}^{C} \phi_{i} \mathcal{N}(\vec{x}|\vec{\mu_{i}},\Sigma_{i})
\end{equation}

$\mathcal{N}$ is a multivariate gaussian defined in \autoref{eq:3}
\begin{equation} \label{eq:3}
	\mathcal{N}(\vec{x}|\vec{\mu_{i}}, \Sigma_{i}) = \frac{1}{\sqrt{(2\pi)^{C}|\Sigma_{i}|}} exp(\tfrac{-1}{2}(\vec{x}-\vec{\mu_{i}})^{T} \Sigma_{i}^{-1}(\vec{x}-\vec{\mu_{i}}))
\end{equation}
where $\vec{x}$ is a data vector. $C$ is the number of components. The $i^{th}$ component parameters are the mean of $\mu_{i}$ and the covariance matrix $\Sigma_{i}$.  The mixture component weigths are defined as $\phi$ with the constraint that $\sum_{i=1}^{C} \phi_{i} = 1$ so that the total probability distribution is normalized to 1.

Selecting the appropriate number of gaussian components was performed by measuring the Bayesian information criterion (BIC) and Akaike information criterion (AIC) for multiple components. The BIC explicited in \autoref{eq:BIC equa} measures the ratio between the likelihood and the number of parameters used, to determine if the likelihood gain is sufficient to justify the number of parameters.
\begin{equation} \label{eq:BIC equa}
	BIC = kln(n) - 2ln(L)
\end{equation}
where $L$ is the maximum value of likelihood, $n$ is the number of data points and $k$ is the number of estimated parameters. Better performing models reduce the BIC indicator. In addition, the AIC evaluates how well a model fits the data it was generated from. The best models according to the AIC are those representing the highest variations while using the fewest independent variables. The AIC is expressed in \autoref{eq:AIC equa} with similar parameters to those in \autoref{eq:BIC equa}. The advantage of these probabilistic model scores is that they do not require test data and can be evaluated on all samples, and handle small datasets.
\begin{equation} \label{eq:AIC equa}
	AIC = 2k - 2ln(L)
\end{equation}

The selected number of gaussians for the GMM is 3, which presented a good trade-off between both BIC and AIC. Maximum values for each indicator were not retained as BIC tends to select too simple models and conversely AIC too complex ones. Values of the probabilistic indicators are summed up in \autoref{tab:BIC and AIC scores} for a varying number of gaussians ranging from 1 to 5.

\begin{table}[H]
\caption{BIC and AIC scores for a varying number of gaussians in GMM, ranging from 1 to 5.}
 \centering{
 \begin{tabular}{|c|c|c|c|c|c|}
 \hline
 \textbf{ $\textbf{n}$ GMM} & 1 & 2 & \textbf{3} & 4 & 5 \\ \hline
 \textbf{BIC} & $-527.40$ & $-532.46$ & $\textbf{-517.70}$ & $-494.46$ & $-482.60$ \\ \hline
 \textbf{AIC} & $-553.01$ & $-586.50$ & $\textbf{-600.17}$ & $-605.39$ & $-621.97$ \\ \hline
 \end{tabular}
}
\label{tab:BIC and AIC scores}
\end{table}
\subsubsection{K-Nearest Neighbors (K-NN)}
K-NN  is a non-parametric model that associates multiple variables by calculating the average of the numerical target of the K-NN. The distance function used for fitting the K-NN model is the Euclidean distance function given in \autoref{eq:4}, as it is widely used and adapted for continuous variable distance measuring.
\begin{equation} \label{eq:4}
	d_{Euclidean}(x, y) = \sqrt{ \sum_{i=1}^{k} (x_{i}-y_{i})^{2} }
\end{equation}
where $x$ and $y$ are data points and $k$ the number of nearest neighbors. To improve performance, the data dimensions are rescaled between 0 and 1, which prevents biasing euclidean distance measures. As the influence of the $k$ value is high, the optimal value is selected to minimize the regression error on the test data split. The error function used to evaluate the performance of the regression is the Root Mean Squared Error (RMSE). This metric was chosen as it keeps the error in the units of the variable of interest. Furthermore, the error is squared before being averaged which penalizes larger errors.~\autoref{eq:RMSE equa} expresses the RMSE error.
\begin{equation} \label{eq:RMSE equa}
	RMSE = \sqrt{\frac{1}{n}\sum_{j=1}^{n} (y_{j} - Y_{j})^{2}}
\end{equation}
with $n$ being the number of data samples, $y$ the predicted target and $Y$ the true label of the target. Overall, the K-NN regression method is well suited to low-dimensional datasets, but loses its practicality as the number of features increases. The optimal value of $k=8$ was retained by performing a grid search cross-validation accross the data and for multiple values of $k$. This was done in order to minimize the prediction error and to limit the bias induced by the data splits during K-NN model fitting.

\subsubsection{ kernel density estimator (KDE)}
KDE is a probability density function estimator for random variables. For each dataset point, it evaluates its probability of belonging to a hypercube. The calculation of the number of points inside the hypercube is formulated in \autoref{eq:5}.
\begin{equation} \label{eq:5}
	k_{n} = \sum_{n=1}^{N} \phi (\dfrac{x-x_{n}}{h})
\end{equation}
where $\phi$ is the window funtion determining whether a dataset entry belongs to the hypercube or not. Knowing the points present inside the hypercube, we can estimate the probability density funtion of the dataset using~\autoref{eq:6}.
\begin{equation} \label{eq:6}
	P(x) = \dfrac{1}{N}\sum_{n=1}^{N} \phi (\dfrac{x-x_{n}}{h})
\end{equation}
with $N$ being the total number of samples, $x$ the center of the hypercube, $x_{n}$ the $n^{th}$ data sample and $h$ the bandwidth of the hypercube. The parameter $h$ has a strong influence on the resulting estimate and must be adapted alongside each data dimension if the data ranges vary. Therefore, we rescaled all our data in range 0 - 1.
For the parsley maturity application, we fitted the KDE to our three-dimensional ground truth dataset composed of (Date, NDVI, CF) to estimate the resultant probability density function of the distribution. The optimal bandwidth selected for our dataset was $h=0.08$ for each dimension. The parameter $h$ was computed by performing a grid search, that is $0.001 \leq h \leq 1.0$,  with a step of $0.001$ and by scoring the KDE fitting on unseen ground truth portions of the dataset.

\subsection{Weak data labeling }
As previously mentioned, we aimed to combine generative models with deep learning approach in order to improve both prediction performance and generalization capabilities of deep networks for CF estimation. The originality of the proposed method lies in the use of generative models to automatically provide weak labels for multispectral images for which no ground truth data exists. Generative labeling introduces bias in the data distribution. This inaccuracy allows for a wider range of potential pairs of CF and multispectral images to be covered, this enabling a better representation of natural fluctuations in the fields. In addition, labeling enough samples of the dataset permits the use of a deep learning approach. The neural network will extract additional features from multispectral images that are not considered during the generative fitting.
\begin{figure*}[t]
	\centerline{
	\includegraphics[width=48.5pc]{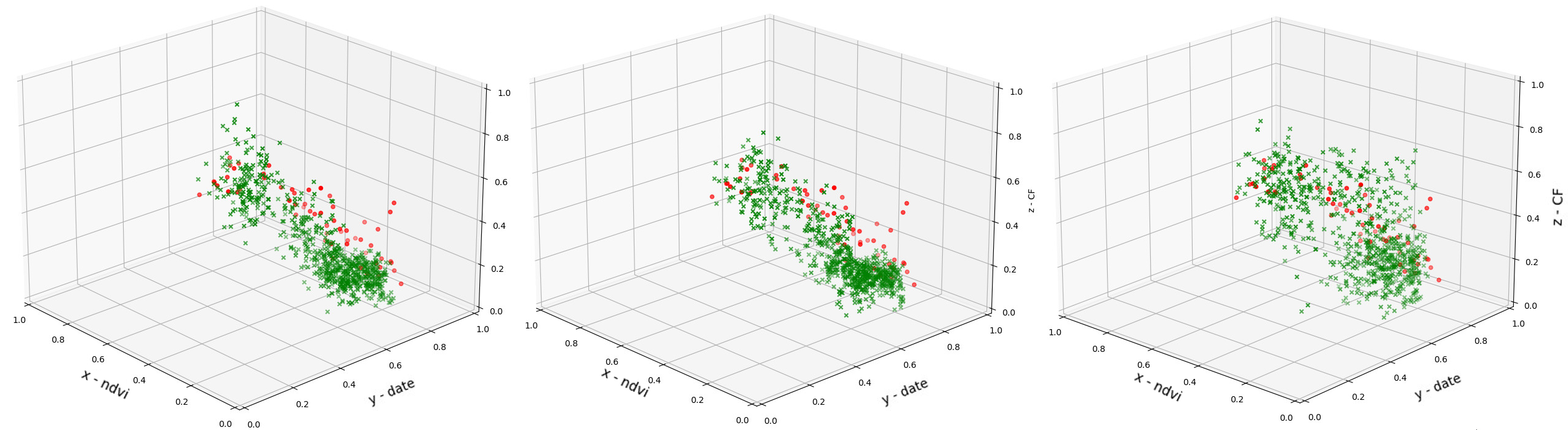}}
	\caption{Generated weak labels in green versus ground truth data in red for GMM (left), K-NN (midle) and KDE (right) for the field A.}
	\label{fig:Generated weak labels in green versus ground truth data in red for GMM, K-NN and KDE}
\end{figure*}

It can be observed in \autoref{fig:Generated weak labels in green versus ground truth data in red for GMM, K-NN and KDE} that the distribution of the generated weak labels varies between the methods, specifically for the KDE model. The labels generated by GMM and K-NN appear compact around the ground truths depicted in red. Therefore, these distributions consider less heterogeneous variations and focus on average variations. Whereas the labels from the KDE span a larger bandwidth of potential data, including the ground truths of the potential heterogeneous data. KDE labeling is seen to be less restricted to the ground truth samples of one field,  but is better suited to generating labels for different fields.

By fixing the variables (Date, NDVI) of the generative model, we can extract a 1D histogram with the possible values of the CF (i.e. the NDVI variable consists of the image mean NDVI). We randomly pick a CF value from the 1D histogram and add the variance to this value in order to extend the range of CF, thus taking into account fluctuations and unseen data. 
The CF obtained is associated with a sample image to constitute its label.  This procedure is applied for all unlabeled NDVI images. 
With these  labeled data, we trained different neural networks in order to improve the CF prediction from NDVI images. 


\subsection{Deep neural networks for CF estimation}
To build a deep learning model for CF estimation, we have chosen two types of popular architectures. The first one is based on convolutional networks (CNN). In this scope, we have opted for popular well performing architectures, namely ResNet and EfficientNet. However, they do not take into account the temporal evolution. For this purpose, we used a second type of architecture, based on Recurrent Neural Networks (RNN), specifically Long Short Term Mermory (LSTM) cells. 
\subsubsection{ResNet}
\begin{figure}[H]
\centerline{\includegraphics[width=12pc]{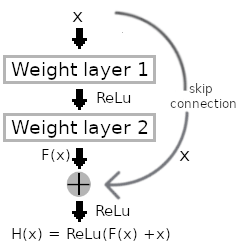}}
\caption{A 2-layer residual block with skip connection.}
\label{fig:A 2-layer residual block with skip connection.}
\end{figure}

Also called deep residual network, it is based on the residual blocks which implement skip connections between layers, as illustrated in \autoref{fig:A 2-layer residual block with skip connection.}. With $\textbf{x}$ the input and $H(\textbf{x})$ the output of each block, the model minimizes the residual function during training, as described in \autoref{eq:residual}.
\begin{equation} \label{eq:residual}
		Residual(\textbf{x}) = H(\textbf{x}) - \textbf{x}
\end{equation} 

The ResNet architecture has shown very good performance in the ImageNet and COCO 2015 competitions. It is implemented in varying depths ranging from 18 to 152 stacked residual blocks. We selected the ResNet-50 version because it incorporates 3-layer residual blocks that perform better than the 2-layer residual blocks used in ResNet-18 and ResNet-34.

\subsubsection{EfficientNet}
As the name of the architecture suggests, the EfficientNet family consists of highly parameter-optimised neural networks. They provide increased parameter accuracy ratio and training efficiency. With only 5.3M parameters for EfficientNetB0, compared to 26 M parameters for ResNet-50, results are slightly better than ResNet-50 on the ImageNet dataset. The neural network optimisation is formulated as an optimisation problem described in \autoref{eq:effnet}, where depth, width and resolution scaling are performed.

\begin{equation} \label{eq:effnet}
	\begin{matrix}
		depth : d = \alpha^{\phi} \\
		width : w = \beta^{\phi} \\
		resolution : r = \gamma^{\phi} \\ 
		
	such\: that : \alpha \times \beta \times \gamma \approx 2 \\
	with : \alpha \geq 1 =, \beta \geq 1, \gamma \geq 1
	\end{matrix}
\end{equation} 
where $\alpha$, $\beta$ and $\gamma$ are constants to dertermine and $\phi$ a coefficient defined by the user.

\subsubsection{LSTM}
\begin{figure}[H]
\centerline{\includegraphics[width=23pc]{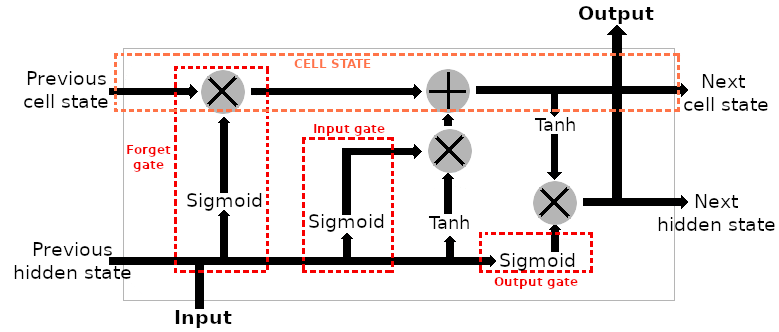}}
\caption{The LSTM cell architecture.}
\label{fig:LSTM cell architecture}
\end{figure}

Long Short Term Memory neural networks are RNNs using LSTM cells. This type of architecture is well suited for sequential data modeling as it considers long-term dependancy between observations and also implements the forget-gate mechanism for discarding irelevant features. The detailed principles of the LSTM cell inner architecture are illustrated in \autoref{fig:LSTM cell architecture} and \autoref{eq:lstm}.
\begin{equation} \label{eq:lstm}
	\begin{matrix}
	i_t = \sigma(W_{i}x_t + U_fh_{t-1}+b_i) \\
	f_t = \sigma(W_{f}x_t + U_fh_{t-1}+b_f) \\
	o_t = \sigma(W_{o}x_t + U_oh_{t-1}+b_o) \\
	c_t = f_t\circ c_{t-1}+i_t\circ \tanh(W_{c}x_t + U_ch_{t-1}+b_c)\\
	h_t = o_t\circ \tanh(c_t)
	\end{matrix}
\end{equation} 
where $W$ and $U$ denote weight terms, $b$ a bias term, and $x_t$ the input. The next hidden state and the previous hidden state are expressed by $h_t$ and $h_{t-1}$ respectively. $c_t$ and $c_{t-1}$ are the states of the next cell and the previous one, respectively. $\sigma(\cdot)$ is the nonlinear Sigmoid activation function. The operator $\circ$ denotes the element-wise product.

As CF maturity may be dependent on both previous and future observations, bidirectional LSTM cells were used, with a 4-sequence length input and a 4-sequence output to quantify each time observation. Since the LSTM input is a sequence vector, a time-distributed CNN head was used as feature extractor on NDVI images  to transform them into the format of the LSTM input.

\section{Experiments and Results}
In this section, we experiment the proposed approaches on real data described previously. First, we evaluate the method of generating labels by parametric and non-parametric techniques. Then, we evaluate the deep learning methods trained with images automatically labeled by these generative methods. To assess performance for each method on unseen parsley fields and varieties, we  conducted  a leave one out cross-validation for each technique. We used rotatively 3 parsley fields as training split for fitting and 1 as test split for evaluation.

\subsection{Evaluation of the generative labeling methods}
The monitored crops  have varying numbers of observations, with 59, 24, 16, and 28 samples from fields A, B, C, and D respectively. Cross-validation procedure for each generative model has been trained using the optimal hyperparameters calculated previously on all  the ground truth data. For instance, we fit GMM, K-NN or KDE on field A, B, C ground truths and evaluate their performance on field D. \autoref{tab:Parametric and Non-parametric train and test split} shows the adopted cross-validation folds, with different sizes of train and test splits.




\begin{table}
\caption{Parametric and Non-parametric train and test split ratios.}
	\centering{
		\begin{tabular}{|c|c|c|c|}
			\hline
			\textbf{Train fields} & \textbf{Train samples} & \textbf{Test field} & \textbf{Test samples} \\ \hline
			B, C, D & $68$  & A & $59$  \\ \hline
			A, C, D & $103$ & B & $24$  \\ \hline
			A, B, D & $111$ & C & $16$ \\ \hline
			A, B, C & $99$  & D & $28$ \\ \hline
		\end{tabular}
	}
	\label{tab:Parametric and Non-parametric train and test split}
\end{table}

As introduced beforehand, GMM, K-NN and KDE were utilized to represent the distribution of the ground truth data samples, for which we have full knowledge of the CF value and the corresponding NDVI images for a given acquisition date. With these models, we firstly evaluate their  prediction capabilities of the CF, and  secondly the best method was used for weak labeling.
\begin{table}
\caption{Root mean square error between true CF and that predicted by GMM, KNN and KDE for each cross-validation fold. }
	\centering{
		\begin{tabular}{|c|ccc|}
			\hline
			Test field &  GMM & K-NN & KDE\\
			\hline
			A & 0.2059 & 0.1418 & \textbf{0.1134} \\ \hline
			B & \textbf{0.1079} & 0.1124 & 0.1084 \\ \hline
			C & 0.0909 & 0.0968 & \textbf{0.0891} \\ \hline
			D & 0.1774 & 0.1782 & \textbf{0.1656} \\ \hline
			Mean & 0.1455 & 0.1323 & \textbf{0.1191} \\ \hline
		\end{tabular}
	}
	\label{tab:K-NN, GMM and KDE Root Mean Squared Error for each Cross-Validation Fold.}
\end{table}
 \autoref{tab:K-NN, GMM and KDE Root Mean Squared Error for each Cross-Validation Fold.}  compares the generative models accuracy on the different test sets with the cross-validation procedure. 
 We can observe that the variability in the number of training samples has a direct impact on the prediction performance of K-NN. The RMSE metric decreases with increasing number of training fold samples, with a maximum RMSE of 0.1782 for the test fold D and a minimum RMSE of 0.0968 for the test fold C. The RMSE of the test field A was not the highest, despite being the largest test set for a model built on the smallest train set. This may be due in part to the fact that the parsley evolution of field A had less steep maturity variations over time and less internal crop disparities. For GMM, we observe similar trends to K-NN for each test fold. However, the changes in RMSE are more proportional to the increase in the number of training samples. Apart from test field A, the performance of GMM compared to K-NN is roughly equivalent, reaching an average error of 0.1422. The  results provided by KDE are better  as observed in~\autoref{tab:K-NN, GMM and KDE Root Mean Squared Error for each Cross-Validation Fold.}. While errors for the folds B and C remain stable for each method, the error of the fold D has decrease by 9\% and fold A reaches a score of 0.1134 compared to 0.1418 or 0.2059 for K-NN and GMM, respectively. On the other hand,  we obtained the best prediction performance with the non-parametric KDE method. This approach seems more robust to variations in the number of training and test samples. Therefore, KDE method was adopted to label image samples. 

To ensure correct quantification of prediction performance, each labeled field was excluded from the training set for KDE model computation. For instance, the field A was labeled using a KDE model fit on fields (B, C, D) in respect to the splits presented previously in \autoref{tab:Parametric and Non-parametric train and test split}. Labeling in this way enables a correct evaluation of the generalization capabilities of the model, since each labeled field is considered as unseen data. As can be seen in \autoref{tab:Weak labeled data for each parsley field} for each field, we now have thousands of weakly labeled data.  Two configurations were considered depending on the deep learning approach used: either a sequence of 4 images for LSTM, or independent images for the CNN-based architecture.

\begin{table}[H]
\caption{Weak labeled data for each parsley field for 32x32 and 128x128 pixel size images per timestep.}
 \centering{
 \begin{tabular}{|c|c|c|c|}
 \hline
 \textbf{Labeled field} & \textbf{32x32} & \textbf{128x128} & \textbf{Num Timesteps} \\ \hline
 A & $64656$ & $4041$ & $15$ \\ \hline
 B & $84208$ & $5263$ & $6$ \\ \hline
 C & $88640$ & $5540$ & $4$ \\ \hline
 D & $73584$ & $4599$ & $7$\\ \hline
 \end{tabular}
}
\label{tab:Weak labeled data for each parsley field}
\end{table}

\subsection{Evaluation of neural regression models}
The weak labeled data was used to train the deep learning models in order to estimate CF from NDVI images. As reported in \autoref{tab:RMSE for each neural networks and labeling models with varying size of input images (32x32 pixels and 128x128 pixels).}, the overall performance of the CNN architectures is similar with a slight advantage for the two EfficientNet models. The size of input images (128$\times$128, 32$\times$32) does not seem to influence result for these architectures either. On the other hand, the LSTM architecture provides the best performance, with a minimum average error of 0.0770, because these models use images from 4 successive observations for CF estimation. Indeed, this type of architecture is better suited for sequential data. An estimation error below 0.1 is equivalent to a 3-4 day variation depending on weather conditions, which highlighs an optimal harvest date for the farmer. Whereas the estimation improvement as compared to the parametric and non-parametric models alone, the neural network models benefit from additional information extracted from multispectral images and enable more specific feature extraction leading to a better estimation. Since the cross-validation procedure was adopted, the prediction capabilities of the models should remain stable for similar parsley fields.

\begin{table*}
\caption{RMSE for each neural networks and labeling models with varying size of input images (32x32 pixels on the left side and 128x128 pixels on the right side of the table).}
\centering
\begin{tabular}{@{}cccc||ccl@{}}
\toprule
     & \textbf{$ResNet50_{32\times32}$} & \textbf{$EfficientNetB0_{32\times32}$}  & \textbf{$LSTM_{32\times32}$} & \textbf{$ResNet50_{128\times128}$} &  \textbf{$EfficientNetB0_{128\times128}$} & \textbf{$LSTM_{128\times128}$} \\ 
\midrule
 	\textbf{GMM} \\
    Field A &  0.1982 &  0.2094 & 0.1782 & 0.1831 & 0.1932 & 0.1743 \\
    Field B &  0.1273 &  0.1249 & 0.1097 & 0.1105 & 0.1056 & 0.1032 \\
    Field C &  0.1129 &  0.1175 & 0.1063 & 0.1074 & 0.1069 & 0.0980 \\
    Field D &  0.1938 &  0.1821 & 0.1632 & 0.1768 & 0.1731 & 0.1564 \\
    Mean	&  0.1581 &  0.1584 & 0.1393 & 0.1445 & 0.1447 & 0.1330 \\ 
  \midrule
 	\textbf{K-NN} \\
    Field A &  0.1564 &  0.1457 & 0.1325 & 0.1435 & 0.1385 & 0.1157 \\
    Field B &  0.1345 &  0.1262 & 0.1141 & 0.1357 & 0.1246 & 0.1083 \\
    Field C &  0.1211 &  0.1147 & 0.1089 & 0.1149 & 0.1051 & 0.0894 \\
    Field D &  0.1718 &  0.1588 & 0.1351 & 0.1522 & 0.1493 & 0.1238 \\
    Mean    &  0.1459 &  0.1363 & 0.1226 & 0.1366 & 0.1294 & 0.1093 \\
  \midrule
    \textbf{KDE} \\
    Field A &  0.1320 &  0.1228 & 0.1133 & 0.1207 & 0.1127 & 0.1021 \\
    Field B &  0.1239 &  0.1136 & 0.0981 & 0.1138 & 0.1047 & 0.0673 \\
    Field C &  0.1152 &  0.1094 & 0.0953 & 0.1042 & 0.0978 & 0.0750 \\
    Field D &  0.1767 &  0.1723 & 0.0902 & 0.1546 & 0.1471 & 0.0636 \\
    Mean 	&  \textbf{0.1366} &  \textbf{0.1295} & \textbf{0.0992} & \textbf{0.1233} & \textbf{0.1156} & \textbf{0.0770} \\

  \bottomrule                          
\end{tabular}
\label{tab:RMSE for each neural networks and labeling models with varying size of input images (32x32 pixels and 128x128 pixels).}
\end{table*}

\section{Discussion}
This study aimed to analyze the possibility of generating extra labeled  data from generative models based on few ground truth samples. Developing such approaches is necessary to enable the creation of datasets large enough for Neural Network applications, especially when applied to agricultural sector. The economic cost and time required to annotate UAV data is high and rarely comprehensive due to continuously changing external factors. To overcome these limitations, as depicted previously, parametric and non-parametric methods have been used to fit ground truths for weak labeling. They were fitted on data components (Date, NDVI, CF) with CF as desired output for chlorophyll concentration quantification.  The scope of this study is limited to these components as they are the most representative to model parsley plant variations, as shown in \autoref{fig:cf and ndvi evolution graphs}.

 Given the small amount of ground truth data, the GMM, K-NN and KDE models performed correctly with a RMSE error varying from 0.2059 to 0.0891. We therefore needed to take this estimation a step further for multiple reasons: ground truths only consider the mean NDVI of images, and ground truths are only few samples from the field. The incorporation of a neural network enabled additional features to be extracted from multispectral images, improving CF estimation, and potentially correcting for manual experimental ground truth sampling errors. Varying the input image size from 32$\times$32 pixels to 128$\times$128 pixels also improved results for LSTM, because instead of having a single plant per image, we have an overview of several plants in a 12 m$^{2}$ area, which better matches the size of the monitored zones.  We also took into account the aspect of temporal variation of the observations, by feeding the recurrent LSTM network with 4 successive observations of the same zone. This was done in order to better address the CF estimation by introducing a factor of vegetation evolution in time.  The dataset being distributed in time, the recurrent neural network models performed better than the CNN models for all folds. An input sequence of 4 observations was used, because the field C was only photographed 4 times before harvest. Longer input sequences should improve the CF estimation.

Finally, in this study, we have showed that large amounts of unlabeled aerial images from UAV can be labeled based on parametric and non-parametric models in order to improve CF estimation and to help generalize neural network prediction on unseen datasets. 
Limitations for future improvement can be highlighted since the acquired data only covers one harvest season and could be subject to weather and/or crop soil type variations.
Acquiring furthermore, aerial images at different heights could also be interesting as parsley types and varieties may have different leaf shapes and reflect light differently. Also more vegetation indices could be combined with NDVI, such as NDRE (Normalized Difference Red-Edge) which is computed from different wavelengths and used for crop nitrogen monitoring. 

\section{Conclusion}
In this paper, we have described a complete workflow for parsley chlorophyll fluorescence (CF) estimation, starting from the aerial UAV image acquisition and agronomic data gathering protocol ending to the CF prediction by neural network model. We have assessed performance of the combination of non-parametric model and neural network approaches to train a general model from few specific ground truth data samples. The KDE yielded the best results for weak labeling of multispectral images and represents higher possible data variations, and recurrent neural networks using LSTM cells achived lower RMSE error compared to CNN architectures. The potential of labeling by generative techniques gives rise to a multitude of possible research directions, especially in remote sensing. In future work, we plan to further improve prediction models and to use more data acquired with different types of modalities, such as satellite images.


\section*{Acknowledgment}

This work was carried out as part of a CIFRE thesis with Frasem company specialised in vegetable seeds multiplication. In was founded by Frasem and Association Nationale de la Recherche et de la Technologie (ANRT).


%





\ifCLASSOPTIONcaptionsoff
  \newpage
\fi





\bibliographystyle{IEEEtran}
\bibliography{references}

\begin{thebibliography}{10}
\providecommand{\url}[1]{#1}
\csname url@samestyle\endcsname
\providecommand{\newblock}{\relax}
\providecommand{\bibinfo}[2]{#2}
\providecommand{\BIBentrySTDinterwordspacing}{\spaceskip=0pt\relax}
\providecommand{\BIBentryALTinterwordstretchfactor}{4}
\providecommand{\BIBentryALTinterwordspacing}{\spaceskip=\fontdimen2\font plus
\BIBentryALTinterwordstretchfactor\fontdimen3\font minus
  \fontdimen4\font\relax}
\providecommand{\BIBforeignlanguage}[2]{{%
\expandafter\ifx\csname l@#1\endcsname\relax
\typeout{** WARNING: IEEEtran.bst: No hyphenation pattern has been}%
\typeout{** loaded for the language `#1'. Using the pattern for}%
\typeout{** the default language instead.}%
\else
\language=\csname l@#1\endcsname
\fi
#2}}
\providecommand{\BIBdecl}{\relax}
\BIBdecl

\bibitem{abd-elmabod_climate_2020}
\BIBentryALTinterwordspacing
S.~K. Abd-Elmabod, M.~Muñoz-Rojas, A.~Jordán, M.~Anaya-Romero, J.~D.
  Phillips, L.~Jones, Z.~Zhang, P.~Pereira, L.~Fleskens, M.~van~der Ploeg, and
  D.~de~la Rosa, ``\BIBforeignlanguage{en}{Climate change impacts on
  agricultural suitability and yield reduction in a {Mediterranean} region},''
  \emph{\BIBforeignlanguage{en}{Geoderma}}, vol. 374, p. 114453, Sep. 2020.
  [Online]. Available:
  \url{https://www.sciencedirect.com/science/article/pii/S0016706119318026}
\BIBentrySTDinterwordspacing

\bibitem{marklein_projected_2020}
\BIBentryALTinterwordspacing
A.~Marklein, E.~Elias, P.~Nico, and K.~Steenwerth,
  ``\BIBforeignlanguage{en}{Projected temperature increases may require shifts
  in the growing season of cool-season crops and the growing locations of
  warm-season crops},'' \emph{\BIBforeignlanguage{en}{Science of The Total
  Environment}}, vol. 746, p. 140918, Dec. 2020. [Online]. Available:
  \url{https://www.sciencedirect.com/science/article/pii/S0048969720344478}
\BIBentrySTDinterwordspacing

\bibitem{maity_climate_2013}
A.~Maity and P.~Pramanik, ``Climate change and seed quality: an alarming issue
  in crop husbandry,'' \emph{Current science}, vol. 105, Nov. 2013.

\bibitem{singh_impacts_2013}
R.~Singh, P.~V.~V. Prasad, and K.~Reddy, ``Impacts of {Changing} {Climate} and
  {Climate} {Variability} on {Seed} {Production} and {Seed} {Industry},''
  \emph{Advances in Agronomy}, vol. 118, pp. 49--110, Jan. 2013.

\bibitem{jalink_werkwijze_1997}
\BIBentryALTinterwordspacing
H.~Jalink, ``Werkwijze voor het bepalen van de rijpheid en kwaliteit van zaden
  middels het chlorofylgehalte en inrichting voor het selecteren van zaden met
  behulp van een dergelijke werkwijze,'' NL Patent NL1\,002\,984C2, Nov., 1997.
  [Online]. Available: \url{https://patents.google.com/patent/NL1002984C2/nl}
\BIBentrySTDinterwordspacing

\bibitem{groot_effect_2006}
S.~Groot, Y.~Birnbaum, N.~Rop, H.~Jalink, G.~Forsberg, C.~Kromphardt,
  S.~Werner, and E.~Koch, ``Effect of seed maturity on sensitivity of seeds
  towards physical sanitation treatments,'' \emph{Seed Science and Technology},
  vol.~34, pp. 403--413, Jul. 2006.

\bibitem{jalink_chlorophyll_1998}
H.~Jalink, R.~v.~d. Schoor, A.~Frandas, J.~G.~v. Pijlen, and R.~J. Bino,
  ``\BIBforeignlanguage{en}{Chlorophyll fluorescence of {Brassica} oleracea
  seeds as a non-destructive marker for seed maturity and seed performance},''
  \emph{\BIBforeignlanguage{en}{Seed Science Research}}, vol.~8, no.~4, pp.
  437--443, Dec. 1998.

\bibitem{jalink_seed_1999}
\BIBentryALTinterwordspacing
H.~Jalink, R.~van~der Schoor, Y.~Birnbaum, and R.~Bino, ``{SEED} {CHLOROPHYLL}
  {CONTENT} {AS} {AN} {INDICATOR} {FOR} {SEED} {MATURITY} {AND} {SEED}
  {QUALITY}.'' \emph{Acta Horticulturae}, no. 504, pp. 219--228, Oct. 1999.
  [Online]. Available: \url{https://www.actahort.org/books/504/504_23.htm}
\BIBentrySTDinterwordspacing

\bibitem{kenanoglu_chlorophyll_2013}
B.~kenanoğlu, ``Chlorophyll fluorescence sorting method to improve quality of
  {Capsicum} pepper seed lots produced from different maturity fruits.''
  \emph{Horticultural Science (HORTSCI)}, vol.~48, Aug. 2013.

\bibitem{demir_chlorophyll_2013}
I.~Demir, B.~Kenanoğlu, H.~Jalink, and K.~Mavi, ``Chlorophyll {Fluorescence}
  {Sorting} {Method} to {Improve} {Seedling} {Emergence} {Potential} and
  {Vigour} of {Commercial} {Tomato} and {Cucumber} {Seed} {Lots},''
  \emph{International Journal of Agriculture and Forestry}, vol.~3, pp.
  333--338, Jan. 2013.

\bibitem{ni_estimation_2019}
Z.~NI, Q.~Lu, H.-Y. Huo, and H.~Zhang, ``Estimation of {Chlorophyll}
  {Fluorescence} at {Different} {Scales}: {A} {Review},'' \emph{Sensors},
  vol.~19, p. 3000, Jul. 2019.

\bibitem{antonio_computerised_2006}
D.~Antonio, ``Computerised seed imaging: {A} new tool to evaluate germination
  quality,'' \emph{Communications in Biometry and Crop Science}, vol.~1, May
  2006.

\bibitem{rodriguez-pulido_preliminary_2012}
\BIBentryALTinterwordspacing
F.~J. Rodríguez-Pulido, R.~Ferrer-Gallego, M.~Lourdes González-Miret, J.~C.
  Rivas-Gonzalo, M.~T. Escribano-Bailón, and F.~J. Heredia,
  ``\BIBforeignlanguage{en}{Preliminary study to determine the phenolic
  maturity stage of grape seeds by computer vision},''
  \emph{\BIBforeignlanguage{en}{Analytica Chimica Acta}}, vol. 732, pp. 78--82,
  Jun. 2012. [Online]. Available:
  \url{https://www.sciencedirect.com/science/article/pii/S0003267012000190}
\BIBentrySTDinterwordspacing

\bibitem{ooms_evaluation_2011}
\BIBentryALTinterwordspacing
D.~Ooms and M.-F. Destain, ``\BIBforeignlanguage{en}{Evaluation of chicory
  seeds maturity by chlorophyll fluorescence imaging},''
  \emph{\BIBforeignlanguage{en}{Biosystems Engineering}}, vol. 110, no.~2, pp.
  168--177, Oct. 2011. [Online]. Available:
  \url{https://www.sciencedirect.com/science/article/pii/S1537511011001279}
\BIBentrySTDinterwordspacing

\bibitem{elmasry_recent_2019}
\BIBentryALTinterwordspacing
G.~ElMasry, N.~Mandour, S.~Al-Rejaie, E.~Belin, and D.~Rousseau,
  ``\BIBforeignlanguage{en}{Recent {Applications} of {Multispectral} {Imaging}
  in {Seed} {Phenotyping} and {Quality} {Monitoring}—{An} {Overview}},''
  \emph{\BIBforeignlanguage{en}{Sensors}}, vol.~19, no.~5, p. 1090, Jan. 2019.
  [Online]. Available: \url{https://www.mdpi.com/1424-8220/19/5/1090}
\BIBentrySTDinterwordspacing

\bibitem{wang_maturity_2021}
\BIBentryALTinterwordspacing
Z.~Wang, X.~Tian, S.~Fan, C.~Zhang, and J.~Li,
  ``\BIBforeignlanguage{en}{Maturity determination of single maize seed by
  using near-infrared hyperspectral imaging coupled with comparative analysis
  of multiple classification models},'' \emph{\BIBforeignlanguage{en}{Infrared
  Physics \& Technology}}, vol. 112, p. 103596, Jan. 2021. [Online]. Available:
  \url{https://www.sciencedirect.com/science/article/pii/S1350449520306447}
\BIBentrySTDinterwordspacing

\bibitem{chakraborty_assessing_2020}
\BIBentryALTinterwordspacing
M.~Chakraborty, L.~R. Khot, and R.~T. Peters,
  ``\BIBforeignlanguage{en}{Assessing suitability of modified center pivot
  irrigation systems in corn production using low altitude aerial imaging
  techniques},'' \emph{\BIBforeignlanguage{en}{Information Processing in
  Agriculture}}, vol.~7, no.~1, pp. 41--49, Mar. 2020. [Online]. Available:
  \url{https://www.sciencedirect.com/science/article/pii/S2214317318302646}
\BIBentrySTDinterwordspacing

\bibitem{modica_monitoring_2020}
\BIBentryALTinterwordspacing
G.~Modica, G.~Messina, G.~De~Luca, V.~Fiozzo, and S.~Praticò,
  ``\BIBforeignlanguage{en}{Monitoring the vegetation vigor in heterogeneous
  citrus and olive orchards. {A} multiscale object-based approach to extract
  trees’ crowns from {UAV} multispectral imagery},''
  \emph{\BIBforeignlanguage{en}{Computers and Electronics in Agriculture}},
  vol. 175, p. 105500, Aug. 2020. [Online]. Available:
  \url{https://www.sciencedirect.com/science/article/pii/S0168169920303045}
\BIBentrySTDinterwordspacing

\bibitem{jiang_uav-based_2021}
\BIBentryALTinterwordspacing
R.~Jiang, A.~Sanchez-Azofeifa, K.~Laakso, P.~Wang, Y.~Xu, Z.~Zhou, X.~Luo,
  Y.~Lan, G.~Zhao, and X.~Chen, ``\BIBforeignlanguage{en}{{UAV}-based partially
  sampling system for rapid {NDVI} mapping in the evaluation of rice nitrogen
  use efficiency},'' \emph{\BIBforeignlanguage{en}{Journal of Cleaner
  Production}}, vol. 289, p. 125705, Mar. 2021. [Online]. Available:
  \url{https://www.sciencedirect.com/science/article/pii/S0959652620357516}
\BIBentrySTDinterwordspacing

\bibitem{shammi_use_2021}
\BIBentryALTinterwordspacing
S.~A. Shammi and Q.~Meng, ``\BIBforeignlanguage{en}{Use time series {NDVI} and
  {EVI} to develop dynamic crop growth metrics for yield modeling},''
  \emph{\BIBforeignlanguage{en}{Ecological Indicators}}, vol. 121, p. 107124,
  Feb. 2021. [Online]. Available:
  \url{https://www.sciencedirect.com/science/article/pii/S1470160X20310633}
\BIBentrySTDinterwordspacing

\bibitem{spadoni_analysis_2020}
\BIBentryALTinterwordspacing
G.~L. Spadoni, A.~Cavalli, L.~Congedo, and M.~Munafò,
  ``\BIBforeignlanguage{en}{Analysis of {Normalized} {Difference} {Vegetation}
  {Index} ({NDVI}) multi-temporal series for the production of forest
  cartography},'' \emph{\BIBforeignlanguage{en}{Remote Sensing Applications:
  Society and Environment}}, vol.~20, p. 100419, Nov. 2020. [Online].
  Available:
  \url{https://www.sciencedirect.com/science/article/pii/S2352938520302858}
\BIBentrySTDinterwordspacing

\bibitem{marques_ramos_random_2020}
\BIBentryALTinterwordspacing
A.~P. Marques~Ramos, L.~Prado~Osco, D.~Elis Garcia~Furuya, W.~Nunes~Gonçalves,
  D.~Cordeiro~Santana, L.~Pereira Ribeiro~Teodoro, C.~Antonio~da Silva~Junior,
  G.~Fernando Capristo-Silva, J.~Li, F.~Henrique Rojo~Baio, J.~Marcato~Junior,
  P.~Eduardo~Teodoro, and H.~Pistori, ``\BIBforeignlanguage{en}{A random forest
  ranking approach to predict yield in maize with uav-based vegetation spectral
  indices},'' \emph{\BIBforeignlanguage{en}{Computers and Electronics in
  Agriculture}}, vol. 178, p. 105791, Nov. 2020. [Online]. Available:
  \url{https://www.sciencedirect.com/science/article/pii/S0168169920319591}
\BIBentrySTDinterwordspacing

\bibitem{yoder_normalized_1994}
\BIBentryALTinterwordspacing
B.~J. Yoder and R.~H. Waring, ``\BIBforeignlanguage{en}{The normalized
  difference vegetation index of small {Douglas}-fir canopies with varying
  chlorophyll concentrations},'' \emph{\BIBforeignlanguage{en}{Remote Sensing
  of Environment}}, vol.~49, no.~1, pp. 81--91, Jul. 1994. [Online]. Available:
  \url{https://www.sciencedirect.com/science/article/pii/0034425794900612}
\BIBentrySTDinterwordspacing

\bibitem{Haboudane2008}
D.~Haboudane, N.~Tremblay, J.~R. Miller, and P.~Vigneault, ``Remote estimation
  of crop chlorophyll content using spectral indices derived from hyperspectral
  data,'' \emph{IEEE Transactions on Geoscience and Remote Sensing}, vol.~46,
  no.~2, pp. 423--437, 2008.

\bibitem{Chaoyang2008}
\BIBentryALTinterwordspacing
C.~Wu, L.~Wang, Z.~Niu, S.~Gao, and M.~Wu, ``Nondestructive estimation of
  canopy chlorophyll content using hyperion and landsat/tm images,''
  \emph{International Journal of Remote Sensing}, vol.~31, no.~8, pp.
  2159--2167, 2010. [Online]. Available:
  \url{https://doi.org/10.1080/01431161003614382}
\BIBentrySTDinterwordspacing

\bibitem{pocas_remote_2020}
\BIBentryALTinterwordspacing
I.~Pôças, A.~Calera, I.~Campos, and M.~Cunha,
  ``\BIBforeignlanguage{en}{Remote sensing for estimating and mapping single
  and basal crop coefficientes: {A} review on spectral vegetation indices
  approaches},'' \emph{\BIBforeignlanguage{en}{Agricultural Water Management}},
  vol. 233, p. 106081, Apr. 2020. [Online]. Available:
  \url{https://www.sciencedirect.com/science/article/pii/S0378377419322498}
\BIBentrySTDinterwordspacing

\bibitem{cao_comparison_2019}
\BIBentryALTinterwordspacing
Z.~Cao, X.~Yao, H.~Liu, B.~Liu, T.~Cheng, Y.~Tian, W.~Cao, and Y.~Zhu,
  ``\BIBforeignlanguage{en}{Comparison of the abilities of vegetation indices
  and photosynthetic parameters to detect heat stress in wheat},''
  \emph{\BIBforeignlanguage{en}{Agricultural and Forest Meteorology}}, vol.
  265, pp. 121--136, Feb. 2019. [Online]. Available:
  \url{https://www.sciencedirect.com/science/article/pii/S0168192318303551}
\BIBentrySTDinterwordspacing

\bibitem{Lang2020}
\BIBentryALTinterwordspacing
L.~Qiao, D.~Gao, J.~Zhang, M.~Li, H.~Sun, and J.~Ma, ``Dynamic influence
  elimination and chlorophyll content diagnosis of maize using uav spectral
  imagery,'' \emph{Remote Sensing}, vol.~12, no.~16, 2020. [Online]. Available:
  \url{https://www.mdpi.com/2072-4292/12/16/2650}
\BIBentrySTDinterwordspacing

\bibitem{tenreiro_using_2021}
\BIBentryALTinterwordspacing
T.~R. Tenreiro, M.~García-Vila, J.~A. Gómez, J.~A. Jiménez-Berni, and
  E.~Fereres, ``\BIBforeignlanguage{en}{Using {NDVI} for the assessment of
  canopy cover in agricultural crops within modelling research},''
  \emph{\BIBforeignlanguage{en}{Computers and Electronics in Agriculture}},
  vol. 182, p. 106038, Mar. 2021. [Online]. Available:
  \url{https://www.sciencedirect.com/science/article/pii/S0168169921000569}
\BIBentrySTDinterwordspacing

\bibitem{penuelas_assessing_1993}
\BIBentryALTinterwordspacing
J.~Peñuelas, J.~A. Gamon, K.~L. Griffin, and C.~B. Field,
  ``\BIBforeignlanguage{en}{Assessing community type, plant biomass, pigment
  composition, and photosynthetic efficiency of aquatic vegetation from
  spectral reflectance},'' \emph{\BIBforeignlanguage{en}{Remote Sensing of
  Environment}}, vol.~46, no.~2, pp. 110--118, Nov. 1993. [Online]. Available:
  \url{https://www.sciencedirect.com/science/article/pii/003442579390088F}
\BIBentrySTDinterwordspacing

\bibitem{lu_evaluating_2017}
\BIBentryALTinterwordspacing
J.~Lu, Y.~Miao, W.~Shi, J.~Li, and F.~Yuan,
  ``\BIBforeignlanguage{en}{Evaluating different approaches to non-destructive
  nitrogen status diagnosis of rice using portable {RapidSCAN} active canopy
  sensor},'' \emph{\BIBforeignlanguage{en}{Scientific Reports}}, vol.~7, no.~1,
  p. 14073, Oct. 2017. [Online]. Available:
  \url{https://www.nature.com/articles/s41598-017-14597-1}
\BIBentrySTDinterwordspacing

\bibitem{elvidge_comparison_1995}
\BIBentryALTinterwordspacing
C.~D. Elvidge and Z.~Chen, ``\BIBforeignlanguage{en}{Comparison of broad-band
  and narrow-band red and near-infrared vegetation indices},''
  \emph{\BIBforeignlanguage{en}{Remote Sensing of Environment}}, vol.~54,
  no.~1, pp. 38--48, Oct. 1995. [Online]. Available:
  \url{https://www.sciencedirect.com/science/article/pii/003442579500132K}
\BIBentrySTDinterwordspacing

\bibitem{Qiao2022}
\BIBentryALTinterwordspacing
L.~Qiao, W.~Tang, D.~Gao, R.~Zhao, L.~An, M.~Li, H.~Sun, and D.~Song,
  ``Uav-based chlorophyll content estimation by evaluating vegetation index
  responses under different crop coverages,'' \emph{Computers and Electronics
  in Agriculture}, vol. 196, p. 106775, 2022. [Online]. Available:
  \url{https://www.sciencedirect.com/science/article/pii/S0168169922000928}
\BIBentrySTDinterwordspacing

\bibitem{Hasan2021}
\BIBentryALTinterwordspacing
A.~S. M.~M. Hasan, F.~Sohel, D.~Diepeveen, H.~Laga, and M.~G. Jones, ``A survey
  of deep learning techniques for weed detection from images,'' \emph{Computers
  and Electronics in Agriculture}, vol. 184, p. 106067, 2021. [Online].
  Available:
  \url{https://www.sciencedirect.com/science/article/pii/S0168169921000855}
\BIBentrySTDinterwordspacing

\bibitem{Ouhami2021}
\BIBentryALTinterwordspacing
M.~Ouhami, A.~Hafiane, Y.~Es-Saady, M.~El~Hajji, and R.~Canals, ``Computer
  vision, iot and data fusion for crop disease detection using machine
  learning: A survey and ongoing research,'' \emph{Remote Sensing}, vol.~13,
  no.~13, 2021. [Online]. Available:
  \url{https://www.mdpi.com/2072-4292/13/13/2486}
\BIBentrySTDinterwordspacing

\bibitem{abiodun_comprehensive_2019}
O.~I. Abiodun, A.~Jantan, A.~E. Omolara, K.~V. Dada, A.~M. Umar, O.~U. Linus,
  H.~Arshad, A.~A. Kazaure, U.~Gana, and M.~U. Kiru, ``Comprehensive {Review}
  of {Artificial} {Neural} {Network} {Applications} to {Pattern}
  {Recognition},'' \emph{IEEE Access}, vol.~7, pp. 158\,820--158\,846, 2019.

\bibitem{ahmed_classification_2012}
\BIBentryALTinterwordspacing
F.~Ahmed, H.~A. Al-Mamun, A.~S. M.~H. Bari, E.~Hossain, and P.~Kwan,
  ``\BIBforeignlanguage{en}{Classification of crops and weeds from digital
  images: {A} support vector machine approach},''
  \emph{\BIBforeignlanguage{en}{Crop Protection}}, vol.~40, pp. 98--104, Oct.
  2012. [Online]. Available:
  \url{https://www.sciencedirect.com/science/article/pii/S026121941200124X}
\BIBentrySTDinterwordspacing

\bibitem{lottes_uav-based_2017}
P.~Lottes, R.~Khanna, J.~Pfeifer, R.~Siegwart, and C.~Stachniss,
  \emph{{UAV}-{Based} {Crop} and {Weed} {Classification} for {Smart}
  {Farming}}, May 2017.

\bibitem{anagnostis_deep_2021}
\BIBentryALTinterwordspacing
A.~Anagnostis, A.~C. Tagarakis, G.~Asiminari, E.~Papageorgiou, D.~Kateris,
  D.~Moshou, and D.~Bochtis, ``\BIBforeignlanguage{en}{A deep learning approach
  for anthracnose infected trees classification in walnut orchards},''
  \emph{\BIBforeignlanguage{en}{Computers and Electronics in Agriculture}},
  vol. 182, p. 105998, Mar. 2021. [Online]. Available:
  \url{https://www.sciencedirect.com/science/article/pii/S0168169921000168}
\BIBentrySTDinterwordspacing

\bibitem{xue_fast_2019}
\BIBentryALTinterwordspacing
W.~Xue, X.~Hu, Z.~Wei, X.~Mei, X.~Chen, and Y.~Xu, ``\BIBforeignlanguage{en}{A
  fast and easy method for predicting agricultural waste compost maturity by
  image-based deep learning},'' \emph{\BIBforeignlanguage{en}{Bioresource
  Technology}}, vol. 290, p. 121761, Oct. 2019. [Online]. Available:
  \url{https://www.sciencedirect.com/science/article/pii/S0960852419309915}
\BIBentrySTDinterwordspacing

\bibitem{tetila_detection_2020}
\BIBentryALTinterwordspacing
E.~C. Tetila, B.~B. Machado, G.~Astolfi, N.~A. d.~S. Belete, W.~P. Amorim,
  A.~R. Roel, and H.~Pistori, ``\BIBforeignlanguage{en}{Detection and
  classification of soybean pests using deep learning with {UAV} images},''
  \emph{\BIBforeignlanguage{en}{Computers and Electronics in Agriculture}},
  vol. 179, p. 105836, Dec. 2020. [Online]. Available:
  \url{https://www.sciencedirect.com/science/article/pii/S016816991831055X}
\BIBentrySTDinterwordspacing

\bibitem{chen_deep_2020}
\BIBentryALTinterwordspacing
H.~Chen, A.~Chen, L.~Xu, H.~Xie, H.~Qiao, Q.~Lin, and K.~Cai,
  ``\BIBforeignlanguage{en}{A deep learning {CNN} architecture applied in smart
  near-infrared analysis of water pollution for agricultural irrigation
  resources},'' \emph{\BIBforeignlanguage{en}{Agricultural Water Management}},
  vol. 240, p. 106303, Oct. 2020. [Online]. Available:
  \url{https://www.sciencedirect.com/science/article/pii/S0378377420307642}
\BIBentrySTDinterwordspacing

\bibitem{kim_regression_2021}
\BIBentryALTinterwordspacing
H.-S. Kim, S.-J. Park, S.-M. Seo, Y.-S. Yoo, H.-W. Jeong, and H.~Jang,
  ``\BIBforeignlanguage{en}{Regression analysis of high-temperature oxidation
  of {Ni}-based superalloys using artificial neural network},''
  \emph{\BIBforeignlanguage{en}{Corrosion Science}}, vol. 180, p. 109207, Mar.
  2021. [Online]. Available:
  \url{https://www.sciencedirect.com/science/article/pii/S0010938X20324884}
\BIBentrySTDinterwordspacing

\bibitem{lathuiliere_comprehensive_2020}
\BIBentryALTinterwordspacing
S.~Lathuilière, P.~Mesejo, X.~Alameda-Pineda, and R.~Horaud, ``A
  {Comprehensive} {Analysis} of {Deep} {Regression},'' \emph{IEEE Transactions
  on Pattern Analysis and Machine Intelligence}, vol.~42, no.~9, pp.
  2065--2081, Sep. 2020, arXiv: 1803.08450. [Online]. Available:
  \url{http://arxiv.org/abs/1803.08450}
\BIBentrySTDinterwordspacing

\bibitem{hamrani_machine_2020}
\BIBentryALTinterwordspacing
A.~Hamrani, A.~Akbarzadeh, and C.~A. Madramootoo,
  ``\BIBforeignlanguage{en}{Machine learning for predicting greenhouse gas
  emissions from agricultural soils},'' \emph{\BIBforeignlanguage{en}{Science
  of The Total Environment}}, vol. 741, p. 140338, Nov. 2020. [Online].
  Available:
  \url{https://www.sciencedirect.com/science/article/pii/S0048969720338602}
\BIBentrySTDinterwordspacing

\bibitem{chen_research_2018}
\BIBentryALTinterwordspacing
S.~Chen, B.~Li, J.~Cao, and B.~Mao, ``\BIBforeignlanguage{en}{Research on
  {Agricultural} {Environment} {Prediction} {Based} on {Deep} {Learning}},''
  \emph{\BIBforeignlanguage{en}{Procedia Computer Science}}, vol. 139, pp.
  33--40, Jan. 2018. [Online]. Available:
  \url{https://www.sciencedirect.com/science/article/pii/S1877050918318830}
\BIBentrySTDinterwordspacing

\bibitem{DBLP:journals/corr/abs-2103-00550}
\BIBentryALTinterwordspacing
X.~Yang, Z.~Song, I.~King, and Z.~Xu, ``A survey on deep semi-supervised
  learning,'' \emph{CoRR}, vol. abs/2103.00550, 2021. [Online]. Available:
  \url{https://arxiv.org/abs/2103.00550}
\BIBentrySTDinterwordspacing

\bibitem{sun_weakly-supervised_2020}
\BIBentryALTinterwordspacing
L.~Sun, P.~Ye, G.~Lyu, S.~Feng, G.~Dai, and H.~Zhang,
  ``\BIBforeignlanguage{en}{Weakly-supervised multi-label learning with noisy
  features and incomplete labels},''
  \emph{\BIBforeignlanguage{en}{Neurocomputing}}, vol. 413, pp. 61--71, Nov.
  2020. [Online]. Available:
  \url{https://www.sciencedirect.com/science/article/pii/S0925231220310900}
\BIBentrySTDinterwordspacing

\bibitem{algan_image_2021}
\BIBentryALTinterwordspacing
G.~Algan and I.~Ulusoy, ``\BIBforeignlanguage{en}{Image classification with
  deep learning in the presence of noisy labels: {A} survey},''
  \emph{\BIBforeignlanguage{en}{Knowledge-Based Systems}}, vol. 215, p. 106771,
  Mar. 2021. [Online]. Available:
  \url{https://www.sciencedirect.com/science/article/pii/S0950705121000344}
\BIBentrySTDinterwordspacing

\bibitem{trentin_parzen_2018}
\BIBentryALTinterwordspacing
E.~Trentin, L.~Lusnig, and F.~Cavalli, ``\BIBforeignlanguage{en}{Parzen neural
  networks: {Fundamentals}, properties, and an application to forensic
  anthropology},'' \emph{\BIBforeignlanguage{en}{Neural Networks}}, vol.~97,
  pp. 137--151, Jan. 2018. [Online]. Available:
  \url{https://www.sciencedirect.com/science/article/pii/S0893608017302332}
\BIBentrySTDinterwordspacing

\bibitem{li_novel_2011}
\BIBentryALTinterwordspacing
B.~Li, Z.~Zhou, D.~Li, and W.~Zou, ``\BIBforeignlanguage{en}{A novel {Parzen}
  probabilistic neural network based noncoherent detection algorithm for
  distributed ultra-wideband sensors},'' \emph{\BIBforeignlanguage{en}{Journal
  of Network and Computer Applications}}, vol.~34, no.~6, pp. 1894--1902, Nov.
  2011. [Online]. Available:
  \url{https://www.sciencedirect.com/science/article/pii/S1084804510002262}
\BIBentrySTDinterwordspacing

\bibitem{nguyen_long-term_2020}
\BIBentryALTinterwordspacing
H.-P. Nguyen, J.~Liu, and E.~Zio, ``\BIBforeignlanguage{en}{A long-term
  prediction approach based on long short-term memory neural networks with
  automatic parameter optimization by {Tree}-structured {Parzen} {Estimator}
  and applied to time-series data of {NPP} steam generators},''
  \emph{\BIBforeignlanguage{en}{Applied Soft Computing}}, vol.~89, p. 106116,
  Apr. 2020. [Online]. Available:
  \url{https://www.sciencedirect.com/science/article/pii/S1568494620300569}
\BIBentrySTDinterwordspacing

\end{thebibliography}
%






\end{document}